\documentclass[runningheads]{llncs}
\usepackage[T1]{fontenc}
\usepackage{graphicx}
\usepackage{booktabs, multirow, color, amsfonts, subfigure}
\usepackage[misc]{ifsym}
\usepackage{hyperref}
\hypersetup{
    colorlinks=true,
    linkcolor=red,
    filecolor=blue,      
    urlcolor=blue,
    citecolor=blue,
}
\begin{document}

\title{SlideGCD: Slide-based Graph Collaborative Training with Knowledge Distillation for Whole Slide Image Classification}
\titlerunning{SlideGCD for Whole Slide Image Classification}

\author{Tong Shu\inst{1} \and
Jun Shi\inst{2}$(\textrm{\Letter})$ \and
Dongdong Sun\inst{1} \and
Zhiguo Jiang\inst{3} \and\\
Yushan Zheng\inst{4}$(\textrm{\Letter})$}
%
\authorrunning{T. Shu et al.}

\institute{School of Computer Science and Information
Engineering, Hefei University of Technology, Hefei 230601, China \and
School of Software, Hefei University of Technology, Hefei 230601, China 
\email{juns@hfut.edu.cn} \and
Image Processing Center, School of Astronautics, Beihang University, Beijing, 102206, China \and
School of Engineering Medicine, Beijing Advanced Innovation Center on Biomedical Engineering, Beihang University, Beijing 100191, China\\
\email{yszheng@buaa.edu.cn}}

\maketitle

\begin{abstract}
Existing WSI analysis methods lie on the consensus that histopathological characteristics of tumors are significant guidance for cancer diagnostics. Particularly, as the evolution of cancers is a continuous process, the correlations and differences across various stages, anatomical locations and patients should be taken into account. However, recent research mainly focuses on the inner-contextual information in a single WSI, ignoring the correlations between slides. To verify whether introducing the slide inter-correlations can bring improvements to WSI representation learning, we propose a generic WSI analysis pipeline SlideGCD that considers the existing multi-instance learning (MIL) methods as the backbone and forge the WSI classification task as a node classification problem. More specifically, SlideGCD declares a node buffer that stores previous slide embeddings for subsequent extensive slide-based graph construction and conducts graph learning to explore the inter-correlations implied in the slide-based graph. Moreover, we frame the MIL classifier and graph learning into two parallel workflows and deploy the knowledge distillation to transfer the differentiable information to the graph neural network. The consistent performance boosting, brought by SlideGCD, of four previous state-of-the-art MIL methods is observed on two TCGA benchmark datasets. The code is available at \href{https://github.com/HFUT-miaLab/SlideGCD}{https://github.com/HFUT-miaLab/SlideGCD}.

\end{abstract}

\section{Introduction}
Histopathology slides contain significant diagnostic information, which leads the slide screening to an indispensable process for cancer diagnoses \cite{HARTMAN20207,zhang2019pathologist}. With the development of computational pathology, computer-aided whole slide image (WSI) analysis is greatly achieved for assisting pathologists in making accurate and reproducible diagnoses efficiently. Owing to hardware limitations and the difficulty of obtaining fine-grained annotation \cite{lu2021data}, the direct process of gigapixel WSIs is nearly impossible or highly expensive. Therefore, computer-aided WSI analysis is usually formulated into the multi-instance learning (MIL) task which considers a WSI as a collection of thousands of patches, analogous to a bag of thousands of instances in the standard multiple instance description. Current patch-based WSI analysis methods focus on how to represent the relationships between patches more comprehensively and efficiently.

The graph-based method is one of the key branches of WSI analysis. Existing graph-based WSI analysis methods follow the same pattern including three stages: 1) patch feature extraction, 2) patch-based graph construction (where one graph describes one slide and nodes represent patches), and 3) graph message passing and graph classification. Compared with the classical MIL method and the sequence-based method, the most significant advantage of the graph-based method lies in the flexibility of the graph construction strategy. Early explorations \cite{chen2021whole,li2018graph,lu2022slidegraph+} directly modeled WSIs as patch-based graphs based on patch similarity in the different spaces. Further research began to involve detailed considerations such as the category of instances \cite{zhang2022whole} and the hierarchical structure of instances \cite{bontempo2023graph,guan2022node}. Recent works began to combine non-simple graphs (e.g. heterogeneous graph \cite{chan2023histopathology} and hypergraph \cite{di2022generating,liang2024caf,shi2024masked}) with morphological characteristics, e.g. lesion state of cell \cite{chan2023histopathology} or tissue topology \cite{di2022generating}, for histopathology WSI analysis. Although there are some works \cite{fan2022cancer,li2023patients,shao2023hvtsurv} have acknowledged the existence of complementary information between the slides from the same patient, the unrestricted slide-level inter-correlations exploration still has not drawn much attention. Since the development of cancers is a continuous process, the correlations and differences across various stages, anatomical locations and patients might also contribute to the slide representation.

\begin{figure}[!t]
    \centering
    \includegraphics[width=0.9\columnwidth]{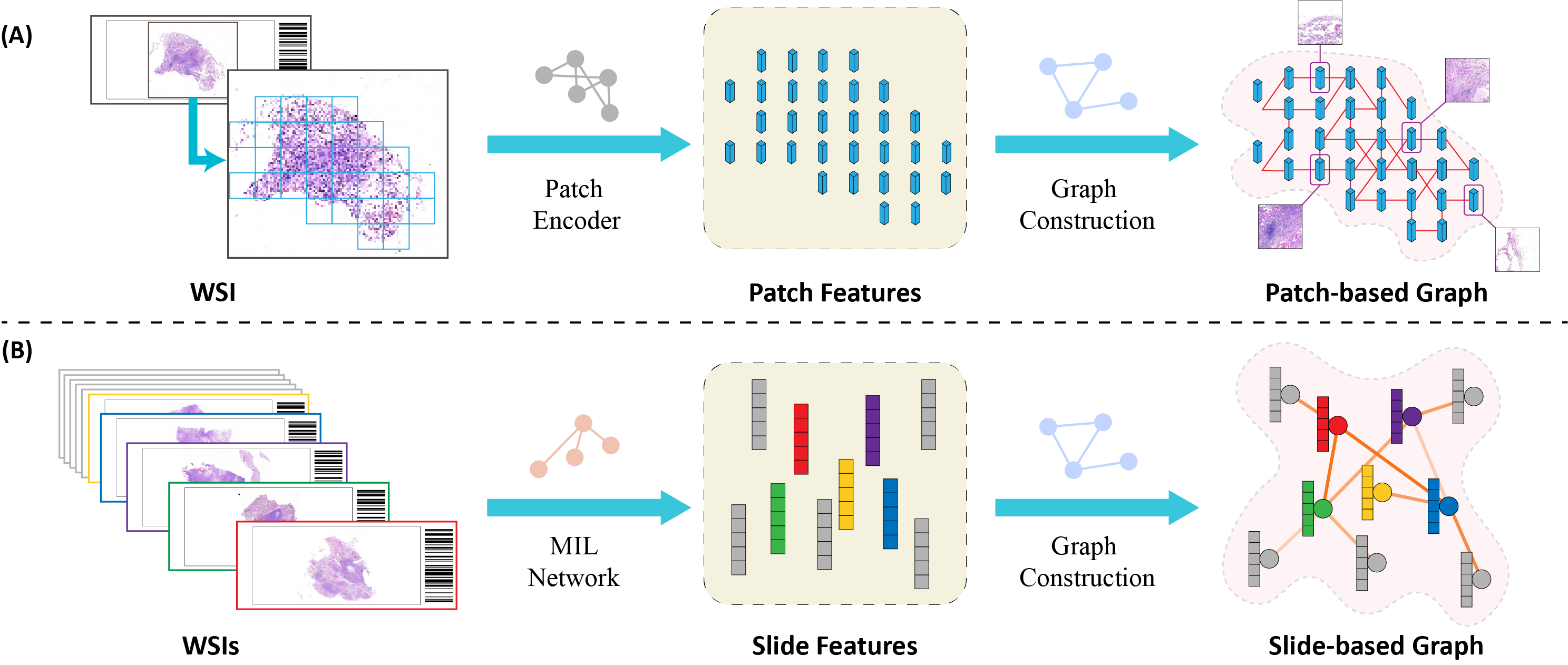}
    \caption{Diagrams of different types of graph in WSI analysis: (A) the Patch-based Graph \& (B) the Slide-based Graph.}
    \label{fig:enter-label}
\end{figure}

Motivated by that consideration, we propose the generic \textbf{Slide}-based \textbf{G}raph \textbf{C}ollaborative training pipeline with knowledge \textbf{D}istillation (SlideGCD) for WSI analysis. The intuitive differences between the patch-based graph and the slide-based graph are shown in Fig. \ref{fig:enter-label}. More specifically, we take existing MIL methods as the backbone of the proposed SlideGCD for obtaining the initial slide embeddings. Then, SlideGCD is used to explore the slide correlations via the slide-based graph. Finally, the slide-level predictions are obtained by node classification. The main contributions of our work are summarized as follows:
\begin{itemize}
\item A generic histopathology WSI analysis pipeline SlideGCD is proposed which can be adapted to any existing MIL methods. This work exploringly frames the WSI dataset into a slide-based graph where WSIs are nodes and mines the slide correlations through graph learning.
\item A rehearsal-based graph construction strategy is deployed to describe the inter-correlations between slides adaptively. Besides, collaborative training for the specially designed GCN with knowledge distillation is applied which aims to fully utilize the well-learned knowledge implied in the MIL classifier.
\end{itemize}

\section{Method}
\subsection{Overview}
\begin{figure}[!t]
    \centering
    \includegraphics[width=0.9\columnwidth]{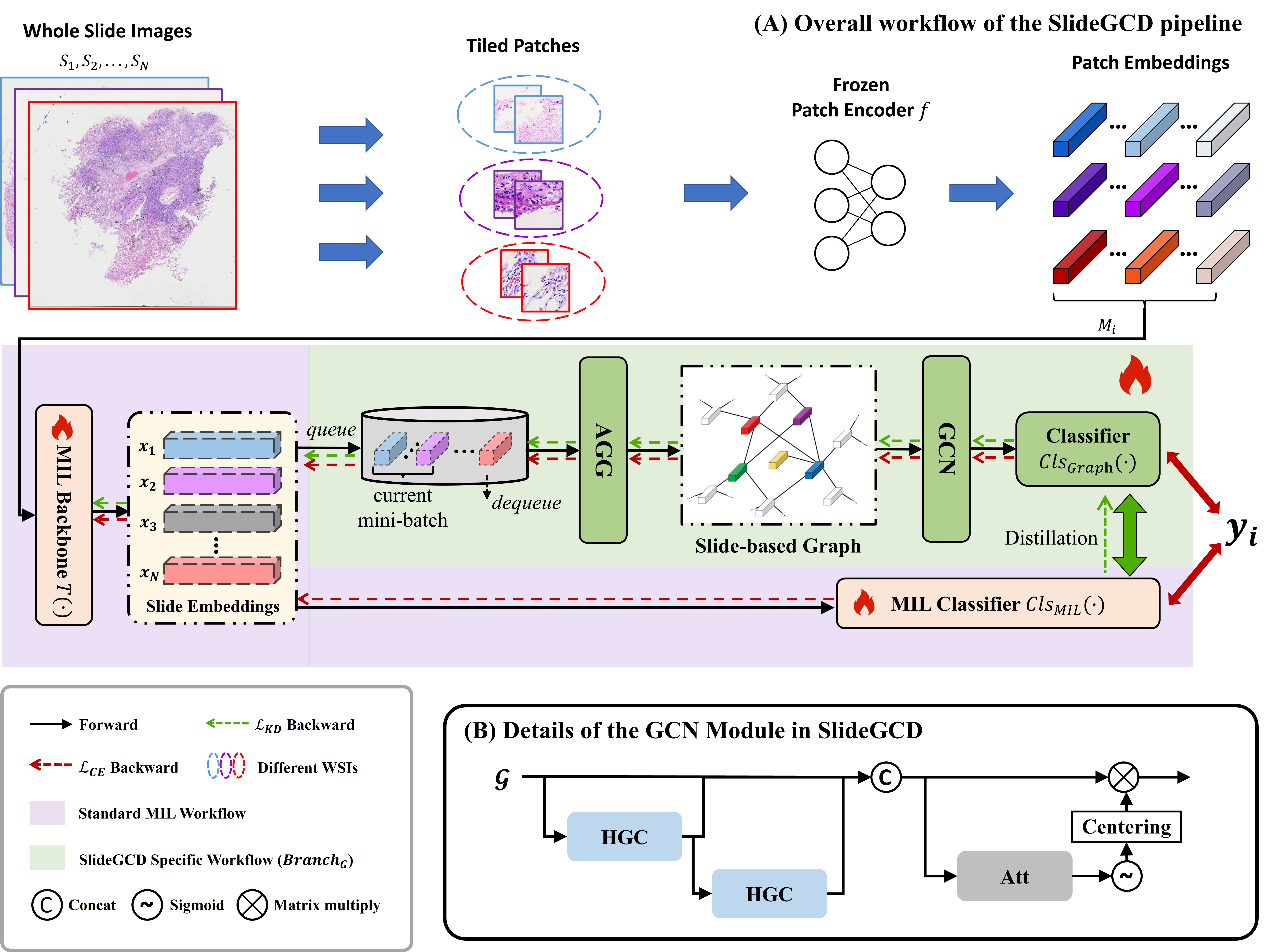}
    \caption{Diagrams of the proposed SlideGCD where (A) describes the overall workflow of the SlideGCD and (B) exhibits the detailed architecture of the GCN in SlideGCD.}
    \label{pipeline}
\end{figure}

The overall workflow of the proposed SlideGCD is shown in Fig. \ref{pipeline}. Similar to the standard MIL framework, SlideGCD first transforms gigapixel WSIs to a series of patch embeddings using a frozen pre-trained patch encoder $f$. Then, a MIL network so-called backbone is deployed to generate the slide embeddings. After that, the slide embeddings are fed into two branches. One branch is actually the rest of the backbone with only the classifier to ensure the convergence and stability of the network during the early training stage. The other graph-based branch is the specific workflow of SlideGCD which explicitly explores the inter-correlations on the extensive slide-based graph.
During the inference stage, the slide-level predictions from the graph-based branch are used as the final predictions.

\subsection{Problem Formulation}
Assuming there is a dataset with $N$ WSIs denoted by $\mathcal{D} = \{(S_i, y_i)\}_{i=1}^N$. Each WSI $S_i = \{p_{i, j}\}_{j=1}^{M_i}$ is annotated with a label $y_i \in \{0, ..., C - 1\}$, where $p_{i, j}$ is tiled patch without patch-level label, $M_i$ is the number of patches and $C$ represents the number of categories. Then, there is a pre-trained patch encoder $f(\cdot)$ to transform the patch $p_k$ into $D_p$ dimensional patch embeddings for reducing the computational cost. The goal of WSI classification is to make prediction $\hat{y_i} = g(f(p_{i, 1}), ..., f(p_{i, M_i}))$ as close as possible to the ground-truth $y_i$, where $g(\cdot)$ is an aggregator that aggregates the instances information and makes final predictions.

\subsection{Backbone: MIL Network}
The proposed SlideGCD aims to explore the inter-correlations between WSIs, and a necessary process is to obtain slide-level representations. We leave this work to the so-called backbone. In an ideal implementation, the backbone $T(\cdot): \mathbb{R}^{N_i \times D_p} \rightarrow \mathbb{R}^{D_S}$ can be any MIL methods with any architecture as long as it can generate fixed dimensional ($D_S$) slide-level embeddings that will be considered the initial node (vertice) embeddings in the subsequent graph-based branch $Branch_{G}$. Notably, we separate the classifier head from the backbone MIL network and denote it as $Cls_{MIL}$ in the following description. The prediction from $Cls_{MIL}$ is denoted as $\hat{y}_{MIL}$.

For generating stable slide embeddings at the training of SlideGCD, there are a few warmup epochs that only optimize the backbone for pre-convergence. Then, the formal training will start with a smaller learning rate, e.g. 1e-4.

\subsection{SlideGCD}
In this section, we will introduce and formulate the details of the specific workflow of SlideGCD. Firstly, a rehearsal-based graph construction module is exploited to generate and update a slide-based graph during training adaptively. Then, a specially designed GCN is deployed to exploit the contextual information from the slide-based graph and refine the slide embeddings. Additionally, a collaborative training strategy with knowledge distillation is employed to solve the problem of knowledge misaligned between $Cls_{MIL}$ and $Branch_{G}$.

\subsubsection{Slide-based Graph Construction.}
Inspired by the idea of Memory Bank \cite{he2020momentum} and the rehearsal buffer \cite{huang2023conslide}, a \textbf{Node Buffer} with a length of $L$ ($L > N$) is designed to store the slide embeddings from the recent training epochs. In each mini-batch training, the current batch of slide embeddings will be pushed into the node buffer and several outdated slide embeddings at the end of the node buffer will be popped out simultaneously (First-in-first-out, like MoCo \cite{he2020momentum}). 
With the node buffer, the involved nodes can be expanded to exceed the capacity of datasets. Specifically, when $L > N$, the node buffer will contain at least two different node embeddings related to the same input slide. The extensive node collection will potentially alleviate the graph heterophily \cite{li2022finding} since similar nodes are more likely to connect.

Considering a general situation where the clinical information of patients is not thoroughly available, we conduct the Adaptive Graph Generation (AGG) strategy to automatically infer the inter-dependencies from the embedding space. Our AGG module consists of two fully connected layers that transform the slide embeddings into an intermediate hidden space. Then, the $k$-NN clustering is conducted and the slides belonging to the same cluster are connected by a hyperedge $\mathcal{E}$, which formulates the slides as a hypergraph $\mathcal{G}$. The adaptive graph generation process can be formulated as: 

\begin{equation}
    \mathcal{G} = (X^{(0)}, \mathcal{E}),\quad \mathcal{E} = KNN(AGG(X^{(0)}), k),
\end{equation}
where $X^{(0)} \in \mathbb{R}^{L \times D_S}$ represents the node embedding sequence in the Node Buffer that contains the current mini-batch data at the head of the sequence. Notably, the nodes retrieved from the buffer are considered static data with no gradients, and gradient propagation only comes from the nodes from the current mini-batch.

\subsubsection{Graph Learning on Slide-based Graph.}
For the slide-based graph $\mathcal{G}$, a designed GCN composed of two hypergraph convolutional layers \cite{bai2021hypergraph} and a Centering-Attention module is applied to explore the context implied in the slide-based graph, as:

\begin{equation}
    X^{(i+1)} = {\rm Leaky\_ReLU}(HGC(X^{(i)}, \mathcal{E})),
\end{equation}
\begin{equation}
    H = Concat(X^{(0)}, X^{(1)}, X^{(2)}), H \in \mathbb{R}^{L \times 3D_S}
\end{equation}
where $HGC(\cdot)$ denotes the hypergraph convolution \cite{bai2021hypergraph} and $X^{(i)}$ contains the information accumulated from the node itself to its $i$-hop neighbors.

To alleviate the graph heterophily, the participation of information from $k$-hop neighbors should be reweighed. Thus, we applied a channel-wise attention module with $Centering$ to rebalance the message and prevent the attention score from always being positive even when facing defective partial information. The computations can be formulated as:

\begin{equation}
    H' = H \cdot Centering(A) = H \cdot (A - Mean(A)),
\end{equation}
\begin{equation}
    A = {\rm Sigmoid}({\rm ReLU}(H^\mathrm{T}W_0)W_1),
\end{equation}
where $W_0, W_1 \in \mathbb{R}^{L \times L}$ are the learnable weights. Finally, an MLP classifier $Cls_{Graph}$ is used to make final predictions for current mini-batch:
\begin{equation}
    \hat{y}_G = {\rm Softmax}({\rm MLP}(H')),
\end{equation}

\subsubsection{Collaborative Training with Knowledge Distillation.}
The network can exploit the slide inter-correlations on the extensive slide-based graph with the above modifications. However, the well-learned intrinsic knowledge of slides implied in the $Cls_{MIL}$ may be neglected. To associate it with the slide inter-correlations and constrain both branches, we involved the knowledge distillation \cite{hinton2015distilling} to transfer the knowledge learned by $Cls_{MIL}$ to the $Branch_G$.

We treat the $Cls_{MIL}$ and the $Branch_G$ as the teacher and student model separately, letting $Branch_{G}$ draw on the beneficial information learned by $Cls_{MIL}$. Specifically, a response-based knowledge distillation loss is adopted as:
\begin{equation}
    L_{KD} = L_{JS}(\hat{y}_G, \hat{y}_{MIL}, \hat{t}),
\end{equation}
where $L_{JS}$ denotes the JS divergence \cite{menendez1997jensen} loss, and $\hat{t}$ is the temperature coefficient.

Then, the final loss of SlideGCD can be written as below, $L_{CE}(\cdot)$ represents the Cross-Entropy loss function, and $w = 1$ is the weight for knowledge distillation:
\begin{equation}
    L = L_{CE}(\hat{y}_{MIL}, Y) + L_{CE}(\hat{y}_G, Y) + w \cdot L_{KD}.
\end{equation}

\section{Experiments}
\subsection{Experimental Setting}
We conducted experiments on two publicly available cancer cohorts (BRCA \& NSCLC) derived from The Cancer Genome Atlas (TCGA) for evaluating the performance of SlideGCD in the WSI classification task. \textbf{TCGA-BRCA} contains 998 diagnostic digital slides of two breast cancer subtypes. Specifically, 794 WSIs of invasive ductal carcinoma (IDC) and 204 WSIs of invasive lobular carcinoma (ILC). \textbf{TCGA-NSCLC} is a collection of two subtype projects for lung cancer, i.e. Lung Squamous Cell Carcinoma (TCGA-LUSC) and Lung Adenocarcinoma (TCGA-LUAD), for a total of 995 diagnostic WSIs, including 496 WSIs of LUSC and 499 WSIs of LUAD. The average accuracy (ACC), macro-average F1 score (F1), and macro-average area under the receiver operating characteristic curve (AUC) are calculated for evaluating the classification performance.

It is worth mentioning that we directly adopted the default setting of hyper-parameters to the baseline from its public repository and remained fixed when applying SlideGCD for the fairness of comparison. More experimental setting details are in Supplementary. During inference, all parameters and the Node Buffer are frozen. When a WSI is inputted, 1) its initial embedding will be made with $T(\cdot)$, 2) the AGG module will insert it into the slide-based graph by connecting it with its k-nearest buffer nodes, 3) the trained GCN will make message passing to refine its embeddings for final classification. We implemented all the models in $PyTorch$ 1.8 and $PyG$ libraries and ran the experiments on a computer with an NVIDIA RTX 3090 GPU.

\begin{table}[!b]
\centering
\caption{The performance of SlideGCD on different baselines. All reported results are the means and standard deviations with five-fold cross-validation (CV). $\dag$: The discrepancies in reported results from the original paper are mainly caused by the difference in patch encoder and CV settings where the original TransMIL conducts a four-fold CV. }
\resizebox{\textwidth}{30mm}{
\begin{tabular}{l|ccc|ccc}
\toprule
\multirow{2}*{Method$\backslash$Metrics}&  \multicolumn{3}{c}{TCGA-BRCA Dataset}& \multicolumn{3}{c}{TCGA-NSCLC Dataset} \\ 
& ACC(\%) & AUC(\%) & F1(\%) & ACC(\%) & AUC(\%) & F1(\%) \\ \midrule
ABMIL \cite{ilse2018attention} (Baseline) 
& 88.97$\pm$0.85 & 89.87$\pm$0.89 & 81.61$\pm$2.27
& 86.96$\pm$1.16 & 95.14$\pm$0.51 & 86.89$\pm$1.20\\
SlideGCD-ABMIL
& 89.04$\pm$1.11 & 90.22$\pm$1.31 & 85.53$\pm$1.53
& 89.57$\pm$0.77 & 95.68$\pm$0.56 & 89.56$\pm$0.78\\ \midrule
Improvement $\Delta$ 
& \textcolor[RGB]{0, 200, 0}{\textbf{+}0.07} & \textcolor[RGB]{0, 200, 0}{\textbf{+}0.45} & \textcolor[RGB]{0, 200, 0}{\textbf{+3.92}}
& \textcolor[RGB]{0, 200, 0}{\textbf{+2.61}} & \textcolor[RGB]{0, 200, 0}{\textbf{+0.54}} & \textcolor[RGB]{0, 200, 0}{\textbf{+2.67}}\\ \midrule

PatchGCN \cite{chen2021whole} (Baseline) 
& 84.80$\pm$1.77 & 87.18$\pm$1.55 & 75.11$\pm$4.57
& 86.62$\pm$2.38 & 94.81$\pm$1.82 & 86.59$\pm$2.43\\
SlideGCD-PatchGCN
& 85.13±0.96 & 87.36±0.96 & 76.08±1.59
& 88.29±1.34 & 95.21±1.08 & 88.26±1.35\\ \midrule
Improvement $\Delta$ 
& \textcolor[RGB]{0, 200, 0}{\textbf{+}{0.33}} & \textcolor[RGB]{0, 200, 0}{\textbf{+}{0.18}} & \textcolor[RGB]{0, 200, 0}{\textbf{+0.97}}
& \textcolor[RGB]{0, 200, 0}{\textbf{+1.67}} & \textcolor[RGB]{0, 200, 0}{\textbf{+}0.40} & \textcolor[RGB]{0, 200, 0}{\textbf{+1.67}}
\\ \midrule

TransMIL$\dag$ \cite{shao2021transmil} (Baseline) 
& 88.17$\pm$1.00& 90.99$\pm$0.91& 82.09$\pm$1.75& 85.82$\pm$1.67& 94.82$\pm$1.17& 85.77$\pm$1.67\\
SlideGCD-TransMIL
& 89.37$\pm$2.23 & 91.14$\pm$1.48 & 82.68$\pm$3.22 
& 86.82$\pm$1.41 & 95.59$\pm$0.27 & 86.78$\pm$1.45\\ \midrule
Improvement $\Delta$ 
& \textcolor[RGB]{0, 200, 0}{\textbf{+1.20}} & \textcolor[RGB]{0, 200, 0}{\textbf{+}0.15} & \textcolor[RGB]{0, 200, 0}{\textbf{+0.59}}
& \textcolor[RGB]{0, 200, 0}{\textbf{+1.00}} & \textcolor[RGB]{0, 200, 0}{\textbf{+0.77}} & \textcolor[RGB]{0, 200, 0}{\textbf{+1.01}}\\ \midrule

DTFDMIL \cite{zhang2022dtfd} (Baseline)
&89.30$\pm$0.44 &90.08$\pm$0.86 & 83.17$\pm$1.43 
&86.42$\pm$1.07 &95.59$\pm$0.66 & 86.40$\pm$1.06 \\
SlideGCD-DTFDMIL
&90.43$\pm$1.65 &91.23$\pm$1.26 & 85.19$\pm$2.59 
&89.83$\pm$1.80 &96.36$\pm$0.51 & 89.82$\pm$1.82\\ \midrule
Improvement $\Delta$ 
& \textcolor[RGB]{0, 200, 0}{\textbf{+1.13}} & \textcolor[RGB]{0, 200, 0}{\textbf{+1.15}} & \textcolor[RGB]{0, 200, 0}{\textbf{+2.02}}
& \textcolor[RGB]{0, 200, 0}{\textbf{+3.41}} & \textcolor[RGB]{0, 200, 0}{\textbf{+0.77}} & \textcolor[RGB]{0, 200, 0}{\textbf{+3.42}}\\ \bottomrule
\end{tabular}}
\label{Comparison with other methods}
\end{table}

\subsection{Effectiveness for WSI Classification}
Table \ref{Comparison with other methods} shows the results for the four previous state-of-the-art MIL methods, namely ABMIL \cite{ilse2018attention}, PatchGCN \cite{chen2021whole}, TransMIL \cite{shao2021transmil} and DTFDMIL \cite{zhang2022dtfd}, with or without the collaboration of SlideGCD. Each of them represents a typical branch in WSI analysis. ABMIL \cite{ilse2018attention} speaks for the classical lightweight attention-based MIL methods without considering patch relationships. PatchGCN \cite{chen2021whole} is a typical graph-based MIL method that involves the patch correlation via the patch-based graph. TransMIL \cite{shao2021transmil} is a powerful transformer-based MIL method that exploits the patch correlations by utilizing the self-attention mechanism. DTFDMIL \cite{zhang2022dtfd} is a novel pseudo-bags-based MIL method that derives the instance probabilities as the attention score of each instance.

Overall, SlideGCN is capable of bringing stable improvement to all four types of baseline. Specifically, on the TCGA-BRCA dataset, SlideGCD-DTFDMIL promotes F1-score from 0.8317 to 0.8519 and improves AUC from 0.9008 to 0.9123. On TCGA-NSCLC dataset, SlideGCD improves DTFDMIL to achieve ACC of 0.8983 \textcolor[RGB]{0, 200, 0}{($\uparrow$3.41\%)}, AUC of 0.9636 \textcolor[RGB]{0, 200, 0}{($\uparrow$0.77\%)} and F1-score of 0.8982 \textcolor[RGB]{0, 200, 0}{($\uparrow$3.42\%)}. Considering the characteristics of datasets and baselines, SlideGCD prefers pseudo-bag-based DTFDMIL most and the performance boosting of lightweight models (i.e. ABMIL\&PatchGCN) on imbalanced dataset TCGA-BRCA is relatively small. Compared with previous methods that only inspect the slide innateness, SlideGCD additionally looks at the potential connection between the WSIs with similar patterns in tumors. Such consideration exploits the WSI dataset more comprehensively and therefore achieves better performance.

As for the extra computations, they mainly come from several additional linear layers and two graph convolutional layers. Concretely, our method has GFLOPs of 404.37 with a mini-batch consisting of 64 WSIs with 5000 patches, which is only 1.01\% higher than the 398.93 GFLOPs of its baseline TransMIL. This increase is minimal.

\begin{figure}[!t]
\centering
\subfigure[]{
    \begin{minipage}[t]{0.33\linewidth}
        \centering
        \includegraphics[width=1\columnwidth]{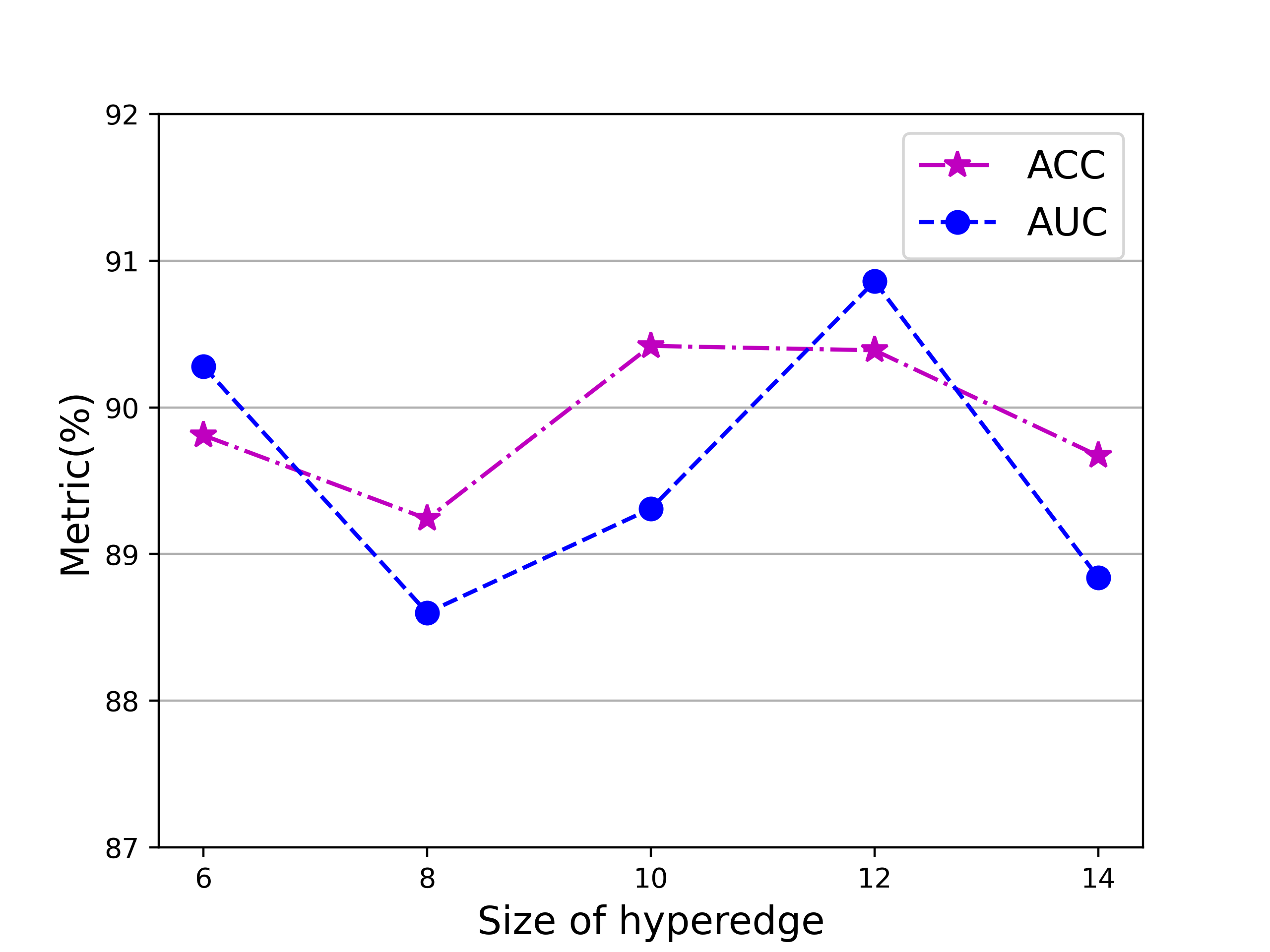}
    \end{minipage}
}%
\subfigure[]{
    \begin{minipage}[t]{0.33\linewidth}
        \centering
        \includegraphics[width=1\columnwidth]{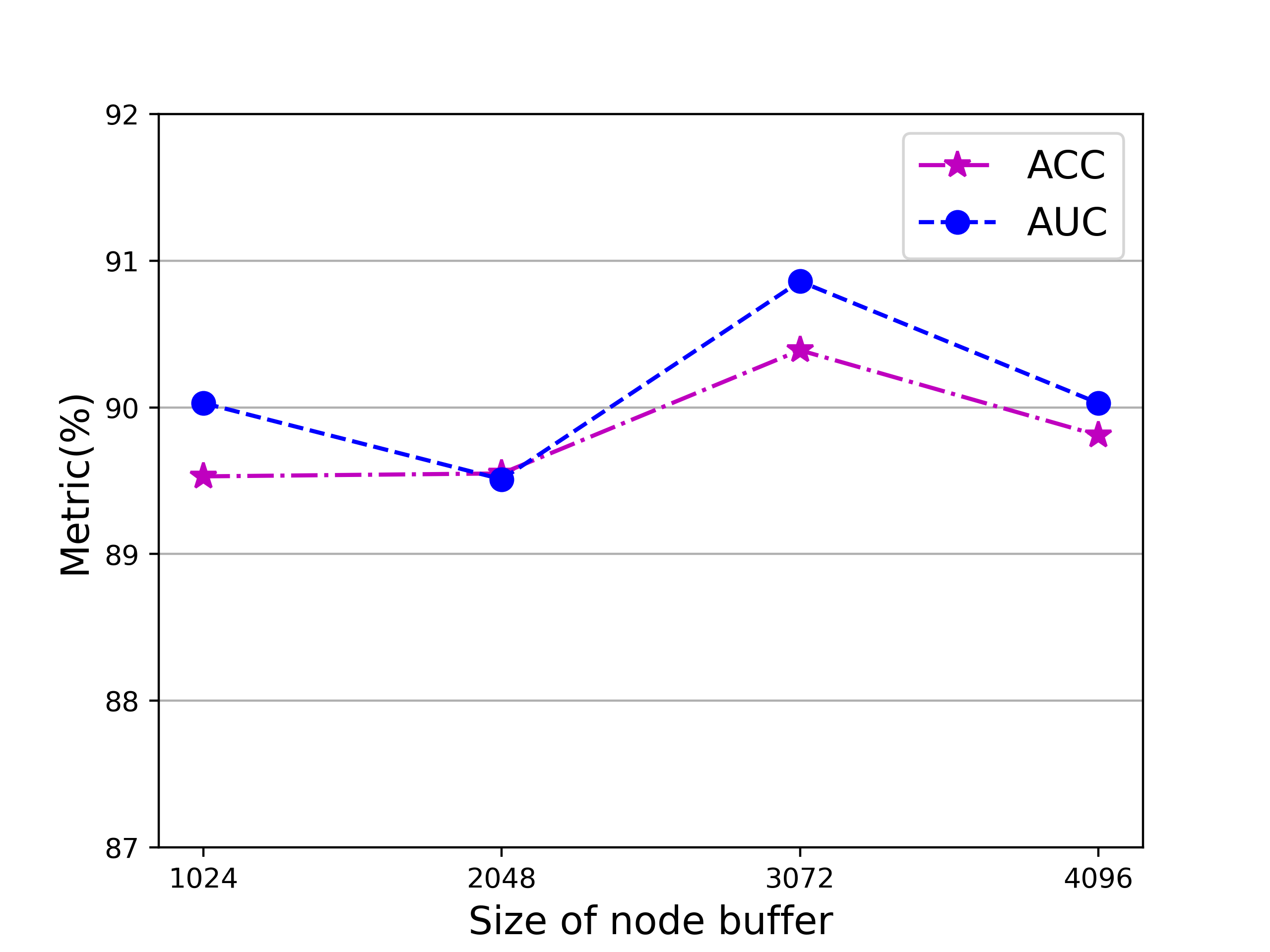}
    \end{minipage}
}%
\subfigure[]{
    \begin{minipage}[t]{0.33\linewidth}
        \centering
        \includegraphics[width=1\columnwidth]{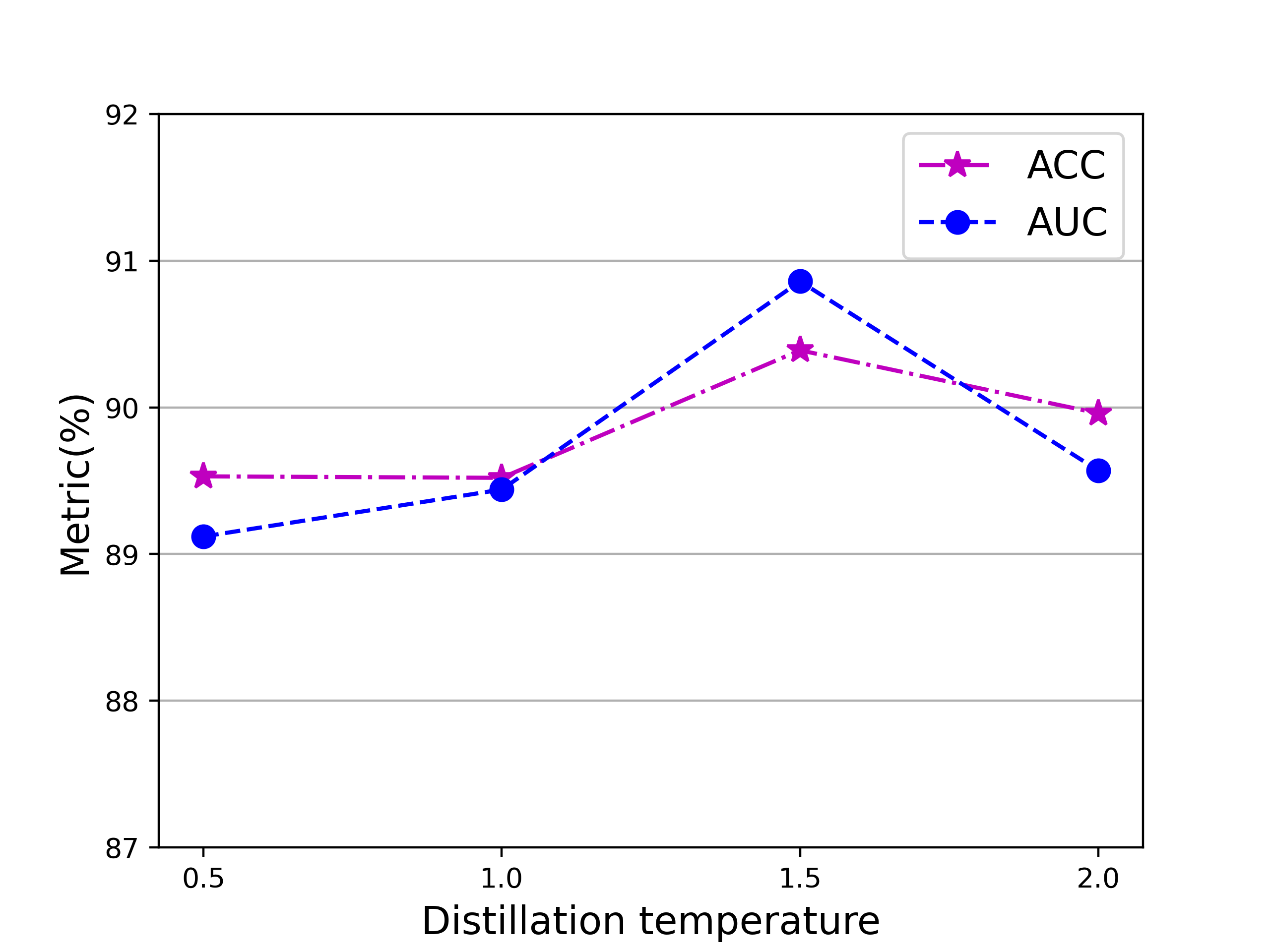}
    \end{minipage}
}%
\caption{Hyperparameter studies of (a) Size of hyperedge $k$, (b) Size of node buffer $L$ and (c) Distillation temperature $\hat{t}$.}
\label{hyperparameters}
\end{figure}

\subsection{Hyperparameter Studies}
Ablation experiments on hyper-parameters are conducted on the TCGA-BRCA dataset with the backbone of DTFDMIL. As presented in Fig. \ref{hyperparameters}, the reported results are the means of ACC$\backslash$AUC on the validation set with five-fold CV. 

\textbf{Size of hyperedge $k$.} $k$ controls the topology of the slide-based graph, and as $k$ increases, there is more overlap between hyperedges. From Fig. \ref{hyperparameters}a, the topology of the slide-based graph could affect the performance of SlideGCD. It does not improve monotonically with the increased size of hyperedge $k$ and seems saturated at $k=12$. Thus, we chose a size of 12 as the default configuration.

\textbf{Size of node buffer $L$.} $L$ affects the timeliness of node features in the constructed slide-based graph. When it becomes too large, it can be foreseen that there will be a decrease in performance as the node buffer contains many outdated slide features. As it becomes too small, the improvement shall descend. According to Fig. \ref{hyperparameters}b, we empirically set $L=3072$.

\textbf{Distillation temperature $\hat{t}$.}  In the higher temperature situation (i.e. $\hat{t}>1$), distillation focuses on transferring knowledge from the teacher model. At lower temperatures (i.e. $\hat{t}<1$), distillation tends to alleviate the impact of noise in negative samples \cite{hinton2015distilling}. Our motivation is to transfer the well-learned knowledge in $Cls_{MIL}$ to $Branch_{G}$, thus a relatively large temperature coefficient should have a better effect. Following the results in Fig. \ref{hyperparameters}c, we set $\hat{t}=1.5$.

\section{Conclusion}
In this paper, we proposed a generic pipeline SlideGCD for histopathology WSI classification and verified the possible improvements of introducing slide correlations via the slide-based graph. SlideGCD exploringly takes the potential connections between slides with similar pathological patterns into account and eventually achieves better performance in WSI classification in a more comprehensive way of utilizing the WSI dataset. Comprehensive experiments have been conducted on various MIL benchmarks and the results show that SlideGCD can boost existing methods consistently.

\begin{credits}
\subsubsection{\ackname} This work was partly supported by the National Natural Science Foundation of China (Grant No. 61906058, 62171007 and 61901018), partly supported by Beijing Natural Science Foundation (Grant No. 7242270), and partly supported by the Fundamental Research Funds for the Central Universities of China (grant No. YWF-23-Q-1075 and JZ2022HGTB0285)

\subsubsection{\discintname}
The authors have no competing interests to declare that are relevant to the content of this article.
\end{credits}

\end{document}